\documentclass[10pt,twocolumn,letterpaper]{article}

\usepackage{cvpr}
\usepackage{times}
\usepackage{epsfig}
\usepackage{graphicx}
\usepackage{amsmath}
\usepackage{amssymb}

\usepackage{caption}
\usepackage{subcaption}
\usepackage{enumitem}
\usepackage{multirow}
\usepackage{booktabs}

\usepackage{amsfonts,amssymb}
\usepackage{graphicx}
\usepackage{algorithm}
\usepackage{algorithmic}

\usepackage{color}
\usepackage[switch]{lineno}
\usepackage{ragged2e}
\usepackage{paralist}
\usepackage{enumitem}
\usepackage{multirow}

\newcommand{\keze}[1]{{\color{black}#1}}

\usepackage[pagebackref=true,breaklinks=true,letterpaper=true,colorlinks,bookmarks=false]{hyperref}

\cvprfinalcopy 

\def\cvprPaperID{1425} 

\ifcvprfinal\fi
\begin{document}

\title{Batch Kalman Normalization: Towards Training Deep Neural Networks \\with Micro-Batches
}

\author{Guangrun Wang$^1$\hspace{28pt}Jiefeng Peng$^1$\hspace{28pt}Ping Luo$^2$\hspace{28pt}Xinjiang Wang$^3$\hspace{28pt}Liang Lin$^{1,3}$\thanks{Corresponding author: Liang Lin (e-mail: linliang@ieee.org). More info about this work can be found at: \url{http://www.sysu-hcp.net}. } \\
\normalsize{$^1$Sun Yat-sen University}\hspace{20pt}
\normalsize{$^2$The Chinese University of Hong Kong}\hspace{20pt}
\normalsize{$^3$SenseTime Group Ltd.}
}

\maketitle

\begin{abstract}
As an indispensable component, Batch Normalization (BN) has successfully improved the training of deep neural networks (DNNs) with mini-batches, by normalizing the distribution of the internal representation for each hidden layer. However, the effectiveness of BN would diminish with scenario of micro-batch (\eg~less than 10 samples in a mini-batch),
since the estimated statistics in a mini-batch are not reliable with insufficient samples.
In this paper, we present a novel normalization method, called Batch Kalman Normalization (BKN), for improving and accelerating the training of DNNs, particularly under the context of micro-batches.
%
%
Specifically, unlike the existing solutions treating each hidden layer as an isolated system, BKN treats all the layers in a network as a whole system, and estimates the statistics of a certain layer by considering the distributions of all its preceding layers, mimicking the merits of Kalman Filtering.
BKN has two appealing properties. First, it enables more stable training and faster convergence compared to previous works.
Second, training DNNs using BKN performs substantially better than those using BN and its variants, especially when very small mini-batches are presented.
{On the image classification benchmark of ImageNet, using BKN powered networks we improve upon the best-published model-zoo results: reaching 74.0\% top-1 \emph{val} accuracy for InceptionV2. More importantly, using BKN achieves the comparable accuracy with extremely smaller batch size, such as 64 times smaller on CIFAR-10/100 and 8 times smaller on ImageNet.}

\end{abstract}

\section{Introduction}

\begin{figure}[t]
\begin{center}
   \includegraphics[width=0.95\linewidth]{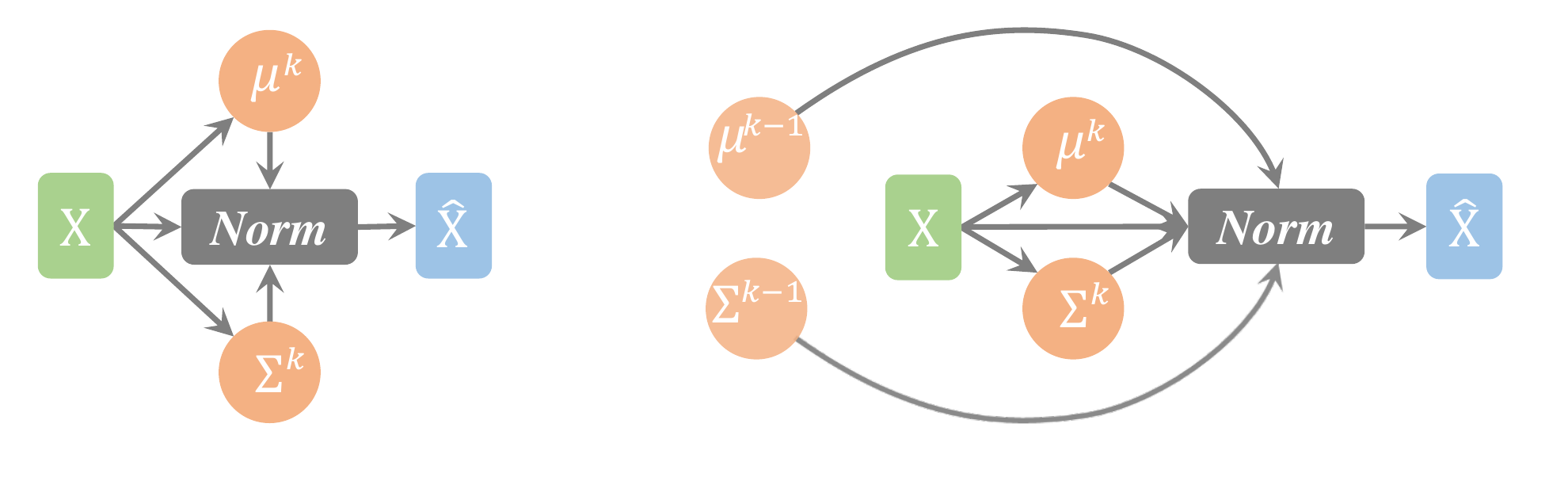}
\end{center}
   \caption{{(a) illustrates the distribution estimation in the conventional Batch Normalization (BN), where the mini-batch statistics, $\mu^k$ and $\Sigma^k$, are estimated based on the currently observed mini-batch at the $k$-th layer.
       For clarity of notation, $\mu^k$ and $\Sigma^k$ indicate the mean and the covariance matrix respectively. Note that only the diagonal entries are used in normalization.
       $X$ and $\widehat{X}$ represent the internal representation before and after normalization.
       In (b), the proposed Batch Kalman Normalization (BKN) provides more accurate distribution estimation of the $k$-th layer, by aggregating the statistics of the preceding ($k$-1)-th layer.
   }}
\label{fig:kalman}
\end{figure}

Batch Normalization (BN) \cite{ioffe2015batch} has recently become a standard and crucial component for improving the training of deep neural networks (DNNs), which is successfully employed to harness several state-of-the-art architectures, such as residual networks \cite{he2016deep} and Inception nets \cite{szegedy2015going}.
In the training and inference of DNNs, BN normalizes the means and variances of the internal representation of each hidden layer,
as illustrated in Figure \ref{fig:kalman} (a).
As pointed out in \cite{ioffe2015batch}, BN enables using larger learning rate in training,
leading to faster convergence.

Although the significance of BN has been demonstrated in many previous works, its drawback cannot be neglected, \ie its effectiveness diminishing when small mini-batch is presented in training.
%
Consider a DNN consisting of a number of layers from bottom to top.
In the traditional BN, the normalization step seeks to eliminate the change in the distributions of its internal layers, by reducing their internal covariant shifts.
Prior to normalizing the distribution of a layer, BN first estimates its statistics, including the means and variances.
However, it is impractical expected for the bottom layer of the input data that can be pre-estimated on the training set, as the representations of the internal layers keep changing after the network parameters have been updated in each training step.
Hence, BN handles this issue by the following schemes. i) During the model training, it approximates the population statistics by using the batch sample statistics in a mini-batch.
ii) It retains the moving average statistics in each training iteration, and employs them during the inference.
However, BN has a limitation, which is limited by the memory capacity of computing platforms (\eg GPUs), especially when the network size and image size are large. In this case, the mini-batch size is not sufficient to approximate the statistics, making them had bias and noise.
And the errors would be amplified when the network becomes deeper, degenerating the quality of the trained model. Negative effects exist also in the inference, where the normalization is applied for each testing sample. Furthermore, in the BN mechanism, the distribution of a certain layer could vary along with the training iteration, which limits the stability of the convergence of the model.

\begin{figure*}[t]
\begin{center}
   \includegraphics[width=0.98\linewidth]{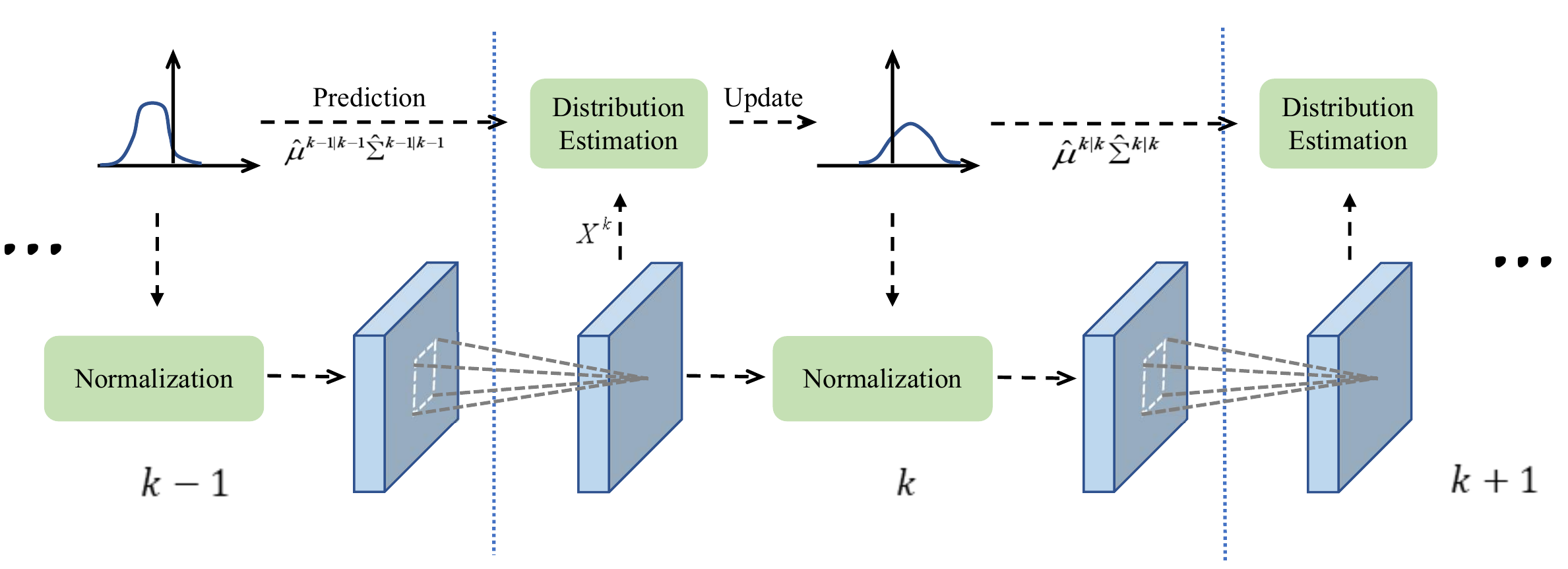}
\end{center}
   \caption{{Illustration of the proposed Batch Kalman Normalization (BKN). At the ($k$-1)-th layer of a DNN, BKN first estimates its statistics (means and covariances), ${\hat{\mu}^{k-1|k - 1}}$ and $\hat{\Sigma}^{k-1|k - 1}$. And the estimations in the $k$-th layer are based on the estimations of the ($k$-1)-th layer, where these estimations are updated by combining with the observed statistics of the $k$-th layer. This process treats the entire DNN as a whole system, different from existing works that estimated the statistics of each hidden layer independently.}}
   \vspace{-3mm}
\label{fig:pipeline}
\end{figure*}

Recently, an extension of BN, called Batch Renormalization (BRN) \cite{ioffe2017batch}, has been proposed to improve the performance of BN when the mini-batch size is small.
During training, BRN performs a correction to the mini-batch statistics by using the moving average statistics, \ie adopting a short-term memory of the mini-batch statistics in the past to make the estimate more robust.
Despite BRN sheds some light on dealing with mini-batches of small sizes, its performance is still undesirable due to the following reasons.
The moving averages are far away from the actual statistics in the early stage of training, making the correction of statistics in BRN unreliable.
In the implementation of BRN, two extra parameters are introduced to measure whether this correction can be trusted, which need to be carefully tuned during training.
Moreover, BRN may probably fail on handling the mini-batches with very few examples, \eg less than 8 samples. In such case, the estimates of the batch sample statistics and moving statistics by either BN or BRN are instable because the means and variances dramatically vary in different training iterations.

In this paper, we present a new normalization method, Batch Kalman Normalization (BKN), for improving and accelerating training of DNNs particularly under the context of micro-batches.
BKN advances the existing solutions by achieving more accurate estimation of the statistics (means and variances) of the internal representations in DNNs. Unlike BN and BRN, where the statistics were estimated by only measuring the mini-batches within a certain layer, \ie they considered each layer in the network as an isolated sub-system, BKN shows that the estimated statistics have strong correlations among the sequential layers.
And the estimations can be more accurately by jointly considering its preceding layers in the network, as illustrated in Figure \ref{fig:kalman} (b).
By analogy, the proposed estimation method shares merits compared to the Kalman filtering process \cite{kalman1960new}.

BKN performs two steps in an iterative way. In the first step, BKN estimates the statistics of the current layer conditioned on the estimations of the previous layer.
In the second step, these estimations are combined with the observed batch sample means and variances calculated within a mini-batch.
These two steps are efficient in BKN.
Updating the current estimation by previous states brings negligible extra computational cost compared to the traditional BN. \keze{Specifically, according to our comprehensive evaluations, the computational complexity of BKN increases only $0.015\times$ compared to that of BN, leading to a marginal additional computation.}

This paper makes the following contributions. 1) We propose an intuitive yet effective  normalization method, offering a promise of improving and accelerating the neural network training. 2) The proposed method enables training networks with mini-batches of very small sizes (\eg~less than $10$ examples), and the resulting models perform substantially better than those using the existing batch normalization methods. This specifically makes our method advantageous in several memory-consuming problems such as training large-scale wide and deep networks, semantic image segmentation, and asynchronous SGD. 3) On the classification benchmark of ImageNet, the experiments show that the recent advanced networks can be strengthened by our method, and the trained models improve the leading results by using less than 60\% training steps \keze{and $0.015\times$ additional computation complexity}.


\section{Related Work}

\textbf{Whitening.}
Decorrelating and whitening the input data \cite{lecun2012efficient} has been demonstrated to speed up training of DNNs.
Some following methods \cite{wiesler2014mean, raiko2012deep, povey2014parallel} were proposed to whiten activations by using sampled training data or performing whitening every thousands iterations to reduce computation.
%
Nevertheless, these operations would lead to model blowing up according to \cite{ioffe2015batch}, because of instability of training.
Recently, the Whitened Neural Network \cite{WNN} and its generalizations \cite{GWNN,EigenNet} presented practical implementations to whiten the internal representation of each hidden layer, and drew the connections between whitened networks and natural gradient descent.
%
Although these approaches had theoretical guarantee and achieved promising results by reducing the computational complexity of the Singular Value Decomposition (SVD) in whitening, their computational costs are still not neglectable, especially when training a DNN with plenty of convolutional layers on a large-scale dataset (\eg ImageNet), as many recent advanced deep architectures did.

\textbf{Standardization.}
To address the above issues, instead of whitening, Ioffe \etal \cite{ioffe2015batch} proposed to normalize the neurons of each hidden layer independently, where the batch normalization (BN) is calculated by using mini-batch statistics.
The extension \cite{cooijmans2016recurrent} adapted BN to recurrent neural networks by using a re-parameterization of LSTM.
In spite of their successes, the heavy dependence of the activations in the entire batch causes some drawbacks to these methods. For example, when the mini-batch size is small, the batch statistics are unreliable.
Hence, several works \cite{salimans2016weight, ba2016layer, arpit2016normalization, salimans2016improved} have been proposed to alleviate the mini-batch dependence. Normalization propagation \cite{arpit2016normalization} attempted to normalize the propagation of the network by using a careful analysis of the nonlinearities, such as the rectified linear units. Layer normalization \cite{ba2016layer} standardized the hidden layer activations, which are invariant to feature shifting and scaling of per training sample. Fixed normalization \cite{salimans2016improved} provided an alternative solution, which employed a separate and fixed mini-batch to compute the normalization parameters.
However,
all of these methods estimated the statistics of the hidden layers separately, whereas BKN treats the entire network as a whole to achieve better estimations.

\section{The Proposed Approach} \label{sec:3}

\textbf{Overview.} Here we introduce some necessary notations that will be used throughout this paper.

Let $x^k$ be the feature vector of a hidden neuron in the $k$-th hidden layer of a DNN, such as a pixel in the hidden convolutional layer of a CNN.
BN normalizes the values of $x^k$ by using a mini-batch of $m$ samples, ${\rm B} = \{ {x_{1}^k,x^k_2,...,x^k_m}\} $.
The mean and covariance of $x^k$ are approximated by ${\bar x^k \leftarrow \frac{1}{m}\sum\limits_{i = 1}^m {{x_i^k}} }$ and ${S^k \leftarrow  {\frac{1}{m}\sum\limits_{i = 1}^m {{{({x_i^k} - \bar x^k)({x_i^k} - \bar x^k)^T}} } } }$.
They are adopted to normalize $x^k$. We have ${{\hat{x}^k} \leftarrow \frac{{{x_i^k} - \bar x^k}}{\sqrt{\mathrm{diag}({S^k})}}}$, where $\mathrm{diag}(\cdot)$ denotes the diagonal entries of a matrix, \ie the variances of $x^k$.
Then, the normalized representation is scaled and shifted to preserve the modeling capacity of the network, ${{y^k} \leftarrow \gamma {{\hat{ x} }^k} + \beta }$, where
$\gamma $ and $\beta $ are parameters that are opmizted in training.
However,
a mini-batch with moderately large size is required to estimate the statistics in BN.
It is compelling to explore better estimations of the distribution in a DNN to accelerate training.

\subsection{Batch Kalman Normalization}

Assume that the true values of the hidden neurons in the $k$-th layer can be represented by the variable $x^k$, which is approximated by using the values in the previous layer $x^{k-1}$. We have
\begin{equation}\label{eq:1}
{{x^k} = {{\bf{A}}^k}{x^{k - 1}} + {u^k}},
\end{equation}
where $\mathbf{A}^k$ is a state transition matrix that transforms the states (features) in the previous layer to the current layer. And $u^k$ is a bias following a Gaussian distribution with zero mean and unit variance.
Note that $\mathbf{A}^k$ could be a linear transition between layers. This is reasonable because our purpose is to 
draw a connection between layers for estimating the statistics rather than accurately compute the hidden features in a certain layer given those in the previous layer.

As the above true values of $x^k$ exist yet not directly accessible, they can be measured by the observation ${z^k}$ with a bias term ${v^k}$,
\begin{equation}\label{eg:kalman}
{{z^k} = {x^k} + {v^k}},
\end{equation}
where $z^k$ indicates the observed values of the features in a mini-batch.
In other words, to estimate the statistics of ${{x^k}}$, previous studies only consider the observed value of ${{z^k}}$ in a mini-batch.
BKN takes into account the features in the previous layer.
To this end, we compute expectation on both sides of Eqn.(\ref{eq:1}), \ie $\mathbb{E}[x^k]=\mathbb{E}[\mathbf{A}^kx^{k-1}+u^k]$, and have
\begin{equation}\label{eq:xestimate}
{{\hat{\mu}^{k|k - 1}} = {{\bf{A}}^k}{\hat{\mu}^{k - 1|k - 1}}},
\end{equation}
where $\hat{\mu}^{k - 1|k - 1}$ denotes the estimation of mean in the ($k$-1)-th layer, and $\hat{\mu}^{k|k - 1}$ is the estimation of mean in the $k$-th layer conditioned on the previous layer.
We call $\hat{\mu}^{k|k - 1}$ an intermediate estimation of the \keze{$k$-th} layer, because it is then combined with the observed values to achieve the final estimation.
As shown in Eqn.(\ref{eq:mu}) below, the estimation in the current layer $\hat{\mu}^{k|k}$ is computed by combining the intermediate estimation with a bias term, which represents the error between the observed values $z^k$ and $\hat{\mu}^{k|k - 1}$. Here $z^k$ indicates the observed mean values and we have $z^k=\overline{x}^k$. And $q^k$ is a gain value indicating how much we reply on this bias.
\begin{equation}
{{\hat{\mu}^{k|k}} = {\hat{\mu}^{k|k - 1}} + {{q}^k}({z^k} - \hat{\mu}^{k|k-1})}.\label{eq:mu}
\end{equation}

Similarly, the estimations of the covariances can be achieved by calculating $\hat{\Sigma}^{k|k-1}=\mathrm{Cov}(x^k-\hat{\mu}^{k|k-1})$ and $\hat{\Sigma}^{k|k}=\mathrm{Cov}(x^k-\hat{\mu}^{k|k})$, where $\mathrm{Cov}(\cdot)$ represents the definition of the covariance matrix.
By introducing $p^k=1 - q ^ k$ and $z^k=\overline{x}^k$, and combining the above definitions with Eqn.(\ref{eq:xestimate}) and (\ref{eq:mu}), we have the following update rules to estimate the statistics as shown in Eqn.(\ref{eq:statistic}). Its proof is given in the Appendix.
\begin{equation}\label{eq:statistic}
\left\{ \begin{array}{l}
{\hat{\mu}^{k|k - 1}} = {{\bf{A}}^k}{\hat{\mu}^{k - 1|k - 1}},\\
{\hat{\mu}^{k|k}} = p^k {\hat{\mu}^{k|k - 1}} + {q^k}{{\bar x}^k},\\
{\hat{\Sigma}^{k|k - 1}} = {{{\bf{A}}}^k}{\hat{\Sigma}^{k - 1|k - 1}}({{{\bf{A}}}^k})^{\rm T} + R,\\
{\hat{\Sigma}^{k|k}} = p^k{\hat{\Sigma}^{k|k - 1}} + {q^k}{S^k}\\
~~~~~~~~~+{p^k}{{q}^k}({{\bar x}^k}{\rm{ - }}{\hat{\mu}^{k|k - 1}})({{\bar x}^k}{\rm{ - }}{\hat{\mu}^{k|k - 1}})^T,
\end{array} \right.
\end{equation}
where ${\hat{\Sigma}^{k|k - 1}}$ and ${\hat{\Sigma}^{k|k}}$ denote the intermediate and the final estimations of the covaraince matrixes in the $k$-th layer respectively. $R$ is the covariance matrix of the bias $u^k$ in Eqn.(\ref{eq:1}). Note that it is identical for all the layers.
$S^k$ are the observed covariance matrix of the mini-batch in the $k$-th layer.
In Eqn.(\ref{eq:statistic}), the transition matrix $\mathbf{A}^k$, the covariance matrix $R$, and the gain value $q^k$ are parameters optimized in training.
In BKN, we employ $\hat{\mu}^{k|k}$ and $\hat{\Sigma}^{k|k}$ to normalize the hidden representation.

\begin{small}
\begin{algorithm}[t]
\caption{Training and Inference wiht Batch Kalman Normalization}
\label{alg:bkn}
\renewcommand{\algorithmicrequire}{ \textbf{Input:}} 
\renewcommand{\algorithmicensure}{ \textbf{Output:}} 
\begin{algorithmic}[1]
\REQUIRE ~~\\
   Values of feature maps $\{ {x_{1...m}}\}$ in the $k^{\rm th}$ layer; statics ${{\hat{ \mu } }^{k-1|k-1}}$ and ${{\hat{ \Sigma } }^{k-1|k-1}}$ in the $(k\mathord{-}1)^{\rm th}$ layer; parameters $\gamma$ and $\beta$; moving mean $\mu$ and moving variance $\Sigma$; moving momentum $\alpha$; Kalman gain $q^k$ and transition matrix ${{{\bf{A}}}^k}$.
\ENSURE ~~\\
   $\{ y_i^k = {\rm{BKN}}( x_i^k ) \}$; updated $\mu$, $\Sigma$; statics ${{\hat{\mu}}^{k|k}}$ and ${{\hat{\Sigma}}^{k|k}}$ in the current layer.
\end{algorithmic}
\renewcommand{\algorithmicrequire}{ \textbf{Train:}} 
\renewcommand{\algorithmicensure}{ \textbf{Inference:}} 
\begin{algorithmic}[1]
\REQUIRE ~~\\
   \begin{equation}
   \begin{aligned}
   &{{\bar x}^k \leftarrow \frac{1}{m}\sum\limits_{i = 1}^m {x_i^k} }\\
   &{S^k \leftarrow  {\frac{1}{m}\sum\limits_{i = 1}^m {{{(x_i^k - {\bar x}^k)(x_i^k - {\bar x}^k)^T}} } } }\\
   &p^k \leftarrow 1 - q ^ k\\
   &{\hat{\mu}^{k|k - 1}} \leftarrow {{\bf{A}}^k}{\hat{\mu}^{k - 1|k - 1}}\\
   &{\hat{\mu}^{k|k}} \leftarrow p^k {\hat{\mu}^{k|k - 1}} + {q^k}{{\bar x}^k}\\
   &{\hat{\Sigma}^{k|k - 1}} \leftarrow {{{\bf{A}}}^k}{\hat{\Sigma}^{k - 1|k - 1}}({{{\bf{A}}}^k})^{\rm T} + R\\
   &{\hat{\Sigma}^{k|k}} \leftarrow p^k{\hat{\Sigma}^{k|k - 1}} \mathord{+} {q^k}{S^k}{\rm{ + }}{p^k}{{q}^k}({{\bar x}^k}{\rm{ - }}{\hat{\mu}^{k|k - 1}})({{\bar x}^k}{\rm{ - }}{\hat{\mu}^{k|k - 1}})^T\\
   &y_i^k \leftarrow \frac{{x_i^k - {\hat{\mu}^{k|k}} }}{\sqrt{\mathrm{diag}(\hat{\Sigma}^{k|k})} }\gamma^k  + \beta^k\\
   &\mathrm{moving~average:}\\
   &\mu := \mu + \alpha (\mu - {\hat{\mu}^{k|k}})\\
   &\Sigma := \Sigma + \alpha (\Sigma - {\hat{\Sigma}^{k|k}})
   \end{aligned}
   \nonumber
   \end{equation}
\ENSURE ~~ ${y_{\rm inference}} \leftarrow \frac{{x - \mu }}{\sqrt{\mathrm{diag}(\Sigma)} }\gamma  + \beta $
\end{algorithmic}
\end{algorithm}
\end{small}

From the above, BKN has two unique characteristics that distinguish it from the BN as well as BRN. First, it offers a better estimation of the distribution. In contrast to the existing normalization methods, the depth information is explicitly exploited in BKN. For instance, the prior message of the distribution of the input image data is leveraged to improve estimation of the second layer's statistics.
On the contrary, ignoring the sequential dependence of the network flow requires larger batch size.
Second, BKN offers a more stable estimation when learning proceeds, where the information flow from prior state to the current state becomes more stable.

Fig.\ref{fig:pipeline} illustrates a diagram of BKN. Unlike BN and BRN, where statistics are computed only within each layer independently, BKN uses messages from all proceeding layers to improve the statistic estimations in the current layer.
Algorithm \ref{alg:bkn} presents Batch Kalman Normalization.
The gradients in SGD for BKN is presented in the Appendix.

\subsection{More Discussions of BKN}\label{sect:small_batch}

In a convolutional layer, different elements of the same feature map at different locations should be normalized in the same way. Therefore, we jointly normalize all the activations in a mini-batch over all locations by following BN. Let ${\rm B}$ be the set of all values in a feature map across both the elements of a mini-batch and spatial locations.
For a mini-batch of size $m$ and the feature maps of spatial size $a \times b$, we have an effective mini-batch of size $m' = \left| {\rm B} \right| = m \times ab$. The parameters $\gamma^{k}$ and $\beta^{k}$ are learned for each feature map, rather than per activation. By taking ResNet101 as an example, the size of the last feature map is $7 \times 7$. When the batch size in training is 32, the sample size for normalization is $7 \times 7 \times 256 = 1568$.

This reveals another benefit of BKN.
The procedure in BKN is a data-driven fusion of the current layer's distribution and the one in previous layer.
This implies that in order to compute the statistics of a certain layer, we achieve them by implicitly using the weighted feature maps of all layers below.
For instance, to compute the mean and variance of the last layer, we are taking into consideration the mean and variance over all the feature maps of the entire network.
More specifically, according to Eqn.(\ref{eq:statistic}), the mean of the last layer $l$ can be computed as ${{\hat{ \mu } }^{l|l}} = {p^l}{{\bf{A}}^l} {{\hat{ \mu } }^{l - 1|l - 1}} + {q^l}{{\bar x}^l}$.
And ${{\hat{ \mu } }^{l - 1|l - 1}}$ can be further decomposed by using the estimations of in the previous ($l$-2) layers.
BKN shares many beneficial properties of both BN and BRN, such as the insensitivity with respect to the initialization of the network parameters and the ability to train efficiently with large learning rate. Unlike BN and BRN, our method ensure that all layers are trained on a reliable estimation, making the training faster and more stable.

\vspace{1mm}
\noindent
{\textbf{Comparison with shortcuts in ResNet.} Although shortcut connection also incorporates information from previous layers, BKN has two unique characteristics that distinguish it from shortcut connection. (1) BKN provides better statistic estimation. In shortcut connection, the informations of previous layer and current layer are simply summed up. No distribution estimation is performed. (2) In theory BKN can be applied to any shortcut connection structure, whereas in practice it may brings memory catastrophe. Because BKN receives only the mean/variance from the previous layer while shortcut connection receives the entire feature maps from the previous layer, the computation cost differs.}

\section{Experiments}\label{sect:exp}

\keze{To demonstrate the generality of our model, we consider two representative networks, i.e., Inceptionv2 \cite{szegedy2015going} and ResNet101 \cite{he2016deep} as the baseline models.}
In \keze{these two} models, BN is stacked after convolution and before the ReLU activation \cite{nair2010rectified}. Thus we denote these two models as ``Inception+BN'' and ``ResNet+BN'', respectively. \keze{Then, we apply our proposed BKN to these two models by simply replacing their employed batch normalization operation, and we denote them as ``Inception+BKN'' and ``ResNet+BKN'', respectively. Similarly, ``Inception+BRN'' / ``ResNet+BRN'' denote replacing BN with the method proposed by BRN~\cite{ioffe2017batch}. }

%

%

We have evaluated \keze{all the methods} on three \keze{image classification} datasets: ImageNet 2012~\cite{russakovsky2015imagenet}, CIFAR-10, and CIFAR-100~\cite{krizhevsky2009learning}.
\keze{As for the ImageNet 2012 benchmark, all the models are trained with {1.28 Million} images and tested on the 50k validation images. We regard the top-1 accuracy as evaluation metric.}

\begin{table}
  \caption{{{\small{Comparison of ImangeNet \emph{val} top-1 accuracy.}}}}
  \scriptsize
  \centering
   \begin{tabular}{l|c c |c}
    \toprule
    & Inceptionv2 & Iterations @ 73.1\% Acc. & ResNet101 \\
    \midrule
    BN &  73.1 &  170k & 77.4 \\
    BKN &  74.0 &  100k & 78.3 \\
    \bottomrule
  \end{tabular}
  \label{tab:baseline}
\end{table}

\subsection{Training with Moderate Batch Size}
\keze{Under this settings, we train Inception+BN and Inception+BKN respectively by using batch size 256 on 8GPUs, i.e., }the mini-batch in each GPU is 32. \keze{Note that,} normalizations are accomplished within each mini-batch, and the gradients are aggregated over 8 GPUs to update the network parameters.

%
%
%
%


Table \ref{tab:baseline} \keze{illustrates the top-1 accuracy on the validation set, where Inception+BKN and Inception+BN} reach 74.0\% and 73.1\% respectively after 170k update steps.
\keze{However}, in terms of reaching 73.1\% accuracy, \keze{Inception+BKN} requires 41.2\% fewer steps than \keze{Inception+BN}.
In particular, Inception+BKN achieves an advanced accuracy of 74.0\% when training converged, outperforming the original network by 1.0\% \cite{ioffe2015batch}.
This improvement is attributed to two reasons.
First, by leveraging the messages from the previous layers, estimation of the statistics is more stable in BKN. \keze{This makes} training converged faster, especially in the early stage.
Second, this procedure also reduces the internal covariance shift, leading to discriminative representation learning and hence improving classification accuracy. {Similar phenomenon can also be observed in ResNet101, where \keze{ResNet+BKN} achieves 78.3\% top-1 accuracy while \keze{ResNet+BN} achieves only 77.4\%.}

\begin{table}[t]
    \caption{\small{\keze{Experiment study on computational complexity analysis}}}
    \scriptsize
    \vspace{-3mm}
    \label{tab:cmp_cost}
    \centering
    \begin{tabular}{c|p{45pt}|p{45pt}}
    \toprule
     & Inception+BN & Inception+BKN  \\
    \midrule
    Speed (examples/sec)& 325.74 & 320.94 \\
    \bottomrule
    \end{tabular}
\end{table}

\vspace{1mm}
\noindent
\textbf{Computation Complexity.} Table \ref{tab:cmp_cost} reports the computation time of Inception+BKN compared to that with Inception+BN, in terms of the number of samples processed per second.
For fair comparison, \keze{all the} methods are individually trained in the same \keze{desktop} with 4 Titan-X GPUs.
\keze{As one can see from Table~\ref{tab:cmp_cost},} Inception+BN and Inception+BKN have similar computational costs, \keze{i.e., our Inception+BKN only performs 0.015$\times$ slower than BN (320.94 vs. 325.74).}


\subsection{Training with Micro-batch}\label{sect:small_batch}
Next we evaluate BKN when batch size is \keze{relatively small under} different settings.
\keze{Specifically,} we define \emph{statistics-batch-size} and \emph{gradient-batch-size} to distinguish these settings.
The former one indicates the sample size used to estimate the statistics in normalization, while the latter one indicates the sample size used to estimate gradients.
Therefore, each setting can be denoted as ``(\emph{gradient-batch-size}, \emph{statistics-batch-size})''.
For example, the above baselines with moderate batch size can be referred as (256,32).

For micro-batch training, its \emph{statistics-batch-size} is typically smaller than 10, leading to non-negligible optimization difficulty.
Training becomes more challenging when the \emph{gradient-batch-size} is also small, \eg smaller than $64$, indicating that not only the statistics of normalization are estimated by using a few samples, but also the gradients are calculated by using a small mini-batch.

\begin{figure}[t]
\centering
\begin{subfigure}{0.5\linewidth}
  \centering
  \includegraphics[width=0.99\linewidth]{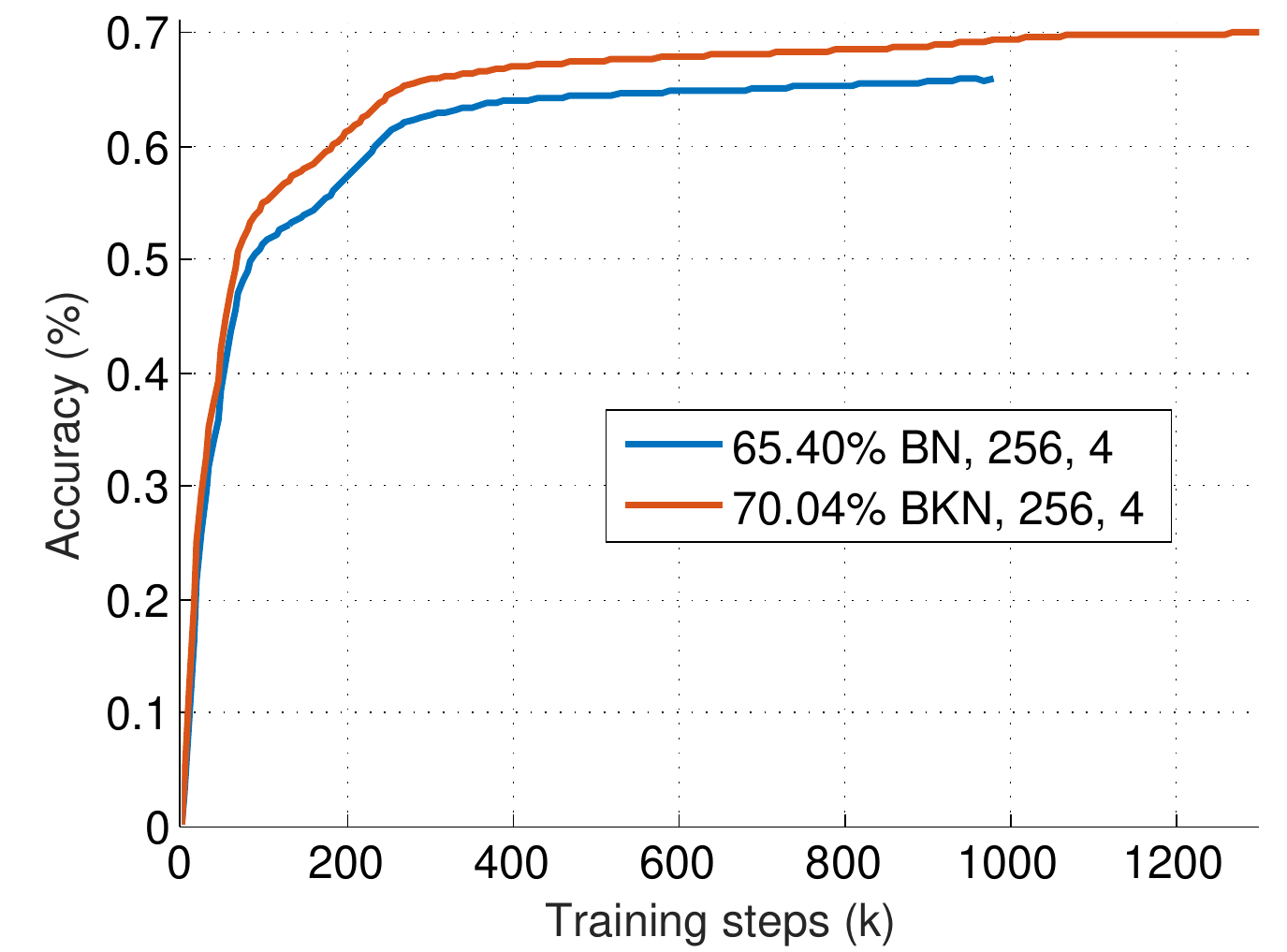}
  \caption{Evaluation on Inceptionv2}
\end{subfigure}%
\begin{subfigure}{0.5\linewidth}
  \centering
  \includegraphics[width=0.99\linewidth]{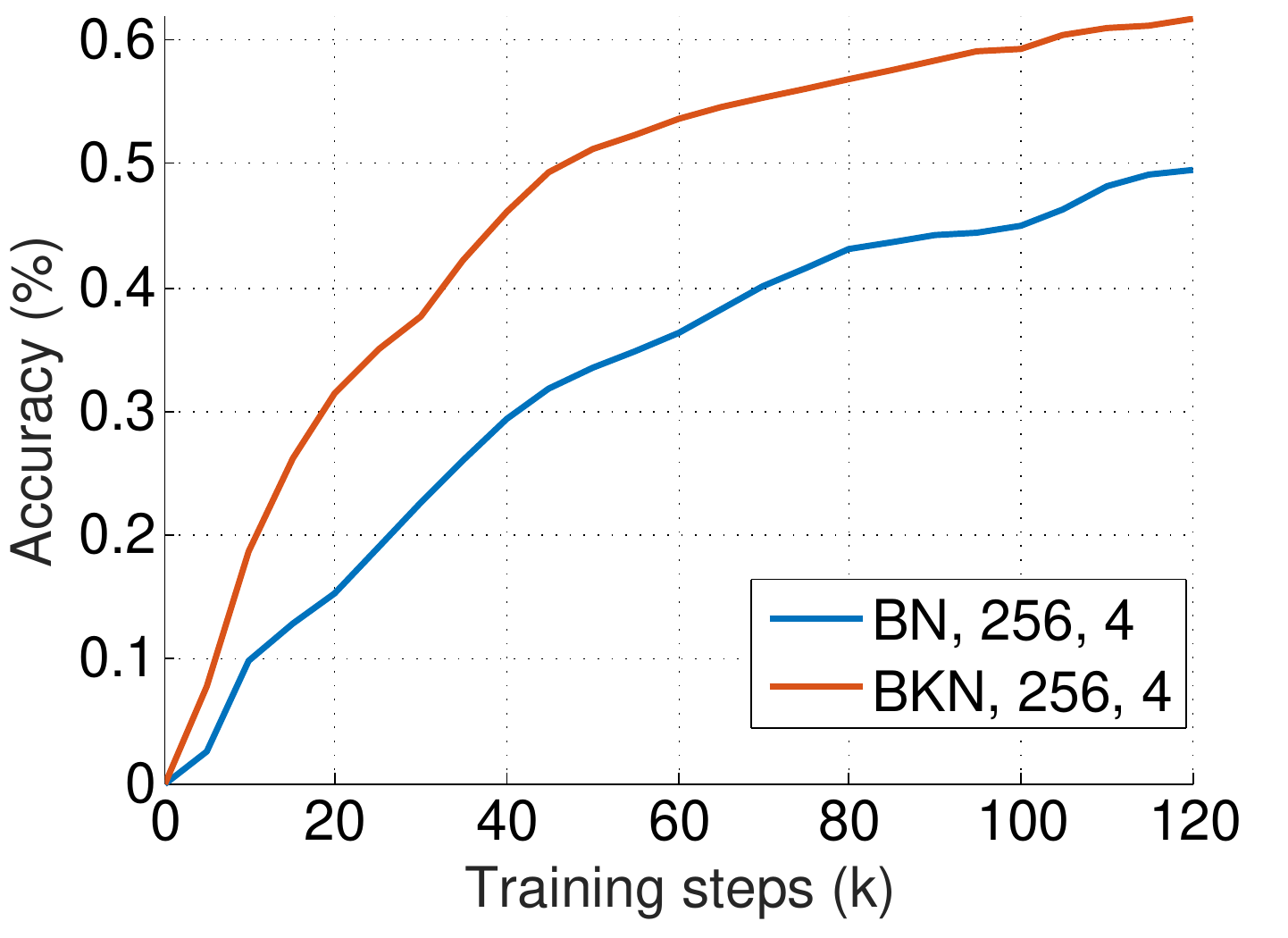}
  \caption{Evaluation on ResNet}
\end{subfigure}
\caption{{Validation accuracy for models trained with BN, and BKN. The setting is (256, 4). BKN allows the model to train faster and achieve a higher accuracy, achieving comparable performance to the baseline setting of (256, 32). (a) presents the evaluation in Inceptionv2, while (b) shows the results of ResNet101. }}
\label{fig:256div4}
\end{figure}

\vspace{3mm}
\noindent
\textbf{Setting of (256,4).}
In this experiment, the \emph{statistic-batch-size} is 4 and the \emph{gradient-batch-size} is 256.
We employ the baseline of (256,32) for comparison.
%
Fig.\ref{fig:256div4} (a) reports the results. We have two major observations from Fig.\ref{fig:256div4}. First, we obtain a substantial improvement of our proposed BKN over BN. For example, in Inceptionv2, Inception+BKN achieves 70.04\% top-1 accuracy, outperforming Inception+BN by a large margin (3.7\%).
%
%
Similar phenomenon can also be observed in ResNet101, for example, at the $120k$-th iteration, ResNet+BKN obtains a top-1 gain of 10.2\% compared to ResNet+BN. These comparisons verify the effectiveness of BKN on micro-batches.

Second, we \keze{have also observed} that under such setting the validation accuracy of both two normalization methods are lower than the baseline \keze{and have a slow training convergence} when facing with \emph{statistic-batch-size} of 32. However, BN is significantly worse compared to the baseline. 
At $1300k$ iterations, Inception+BKN achieves 70.04\%, which is comparable to the baseline (73.1\%) by using 8 times smaller batch size. This indicates that the micro-batch training problem is better addressed by using BKN \keze{rather than} BN.

\begin{figure}[t]
\centering
\begin{subfigure}{0.52\linewidth}
  \centering
  \includegraphics[width=0.99\linewidth]{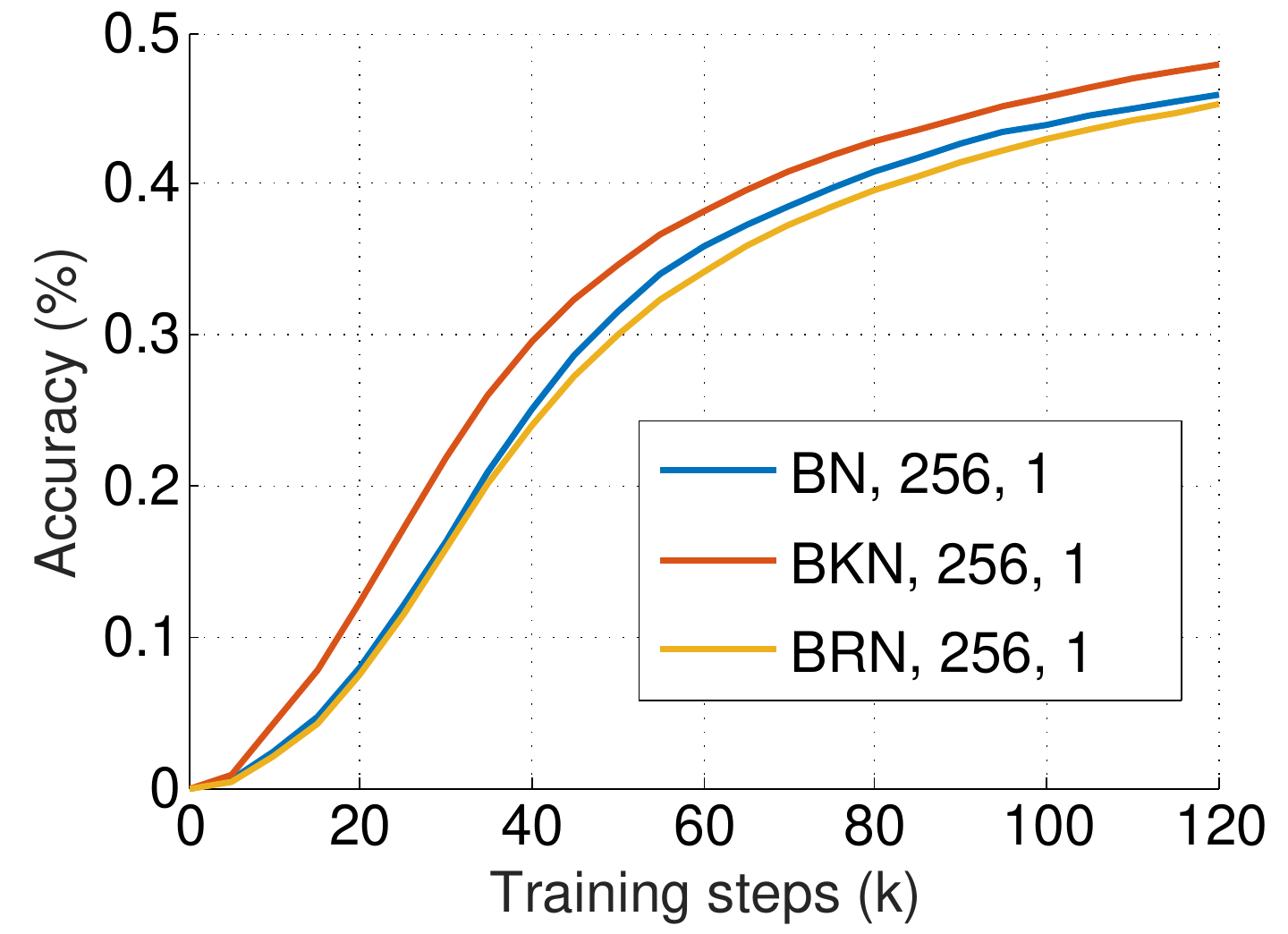}
  \caption{Using batch sample statistics}
\end{subfigure}%
\begin{subfigure}{0.47\linewidth}
  \centering
  \includegraphics[width=0.99\linewidth]{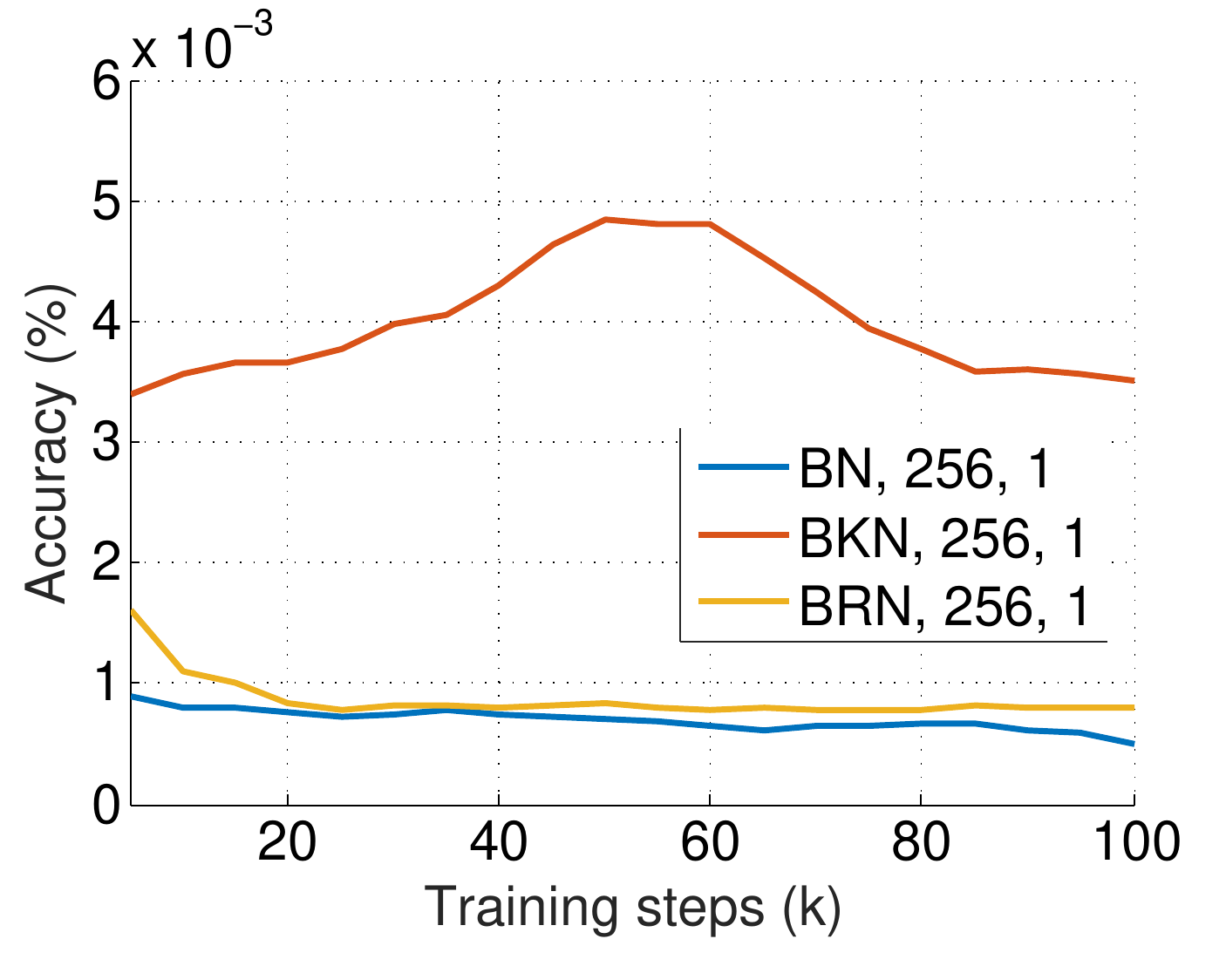}
  \caption{Using population statistics}
\end{subfigure}
\caption{{Validation accuracy for models trained with BN, BRN, and BKN, where the setting is (256, 1). BKN achieves a higher accuracy.}}
\label{fig:256div1}
\end{figure}

\begin{figure}[t]
\centering
\begin{subfigure}{0.5\linewidth}
  \centering
  \includegraphics[width=0.99\linewidth]{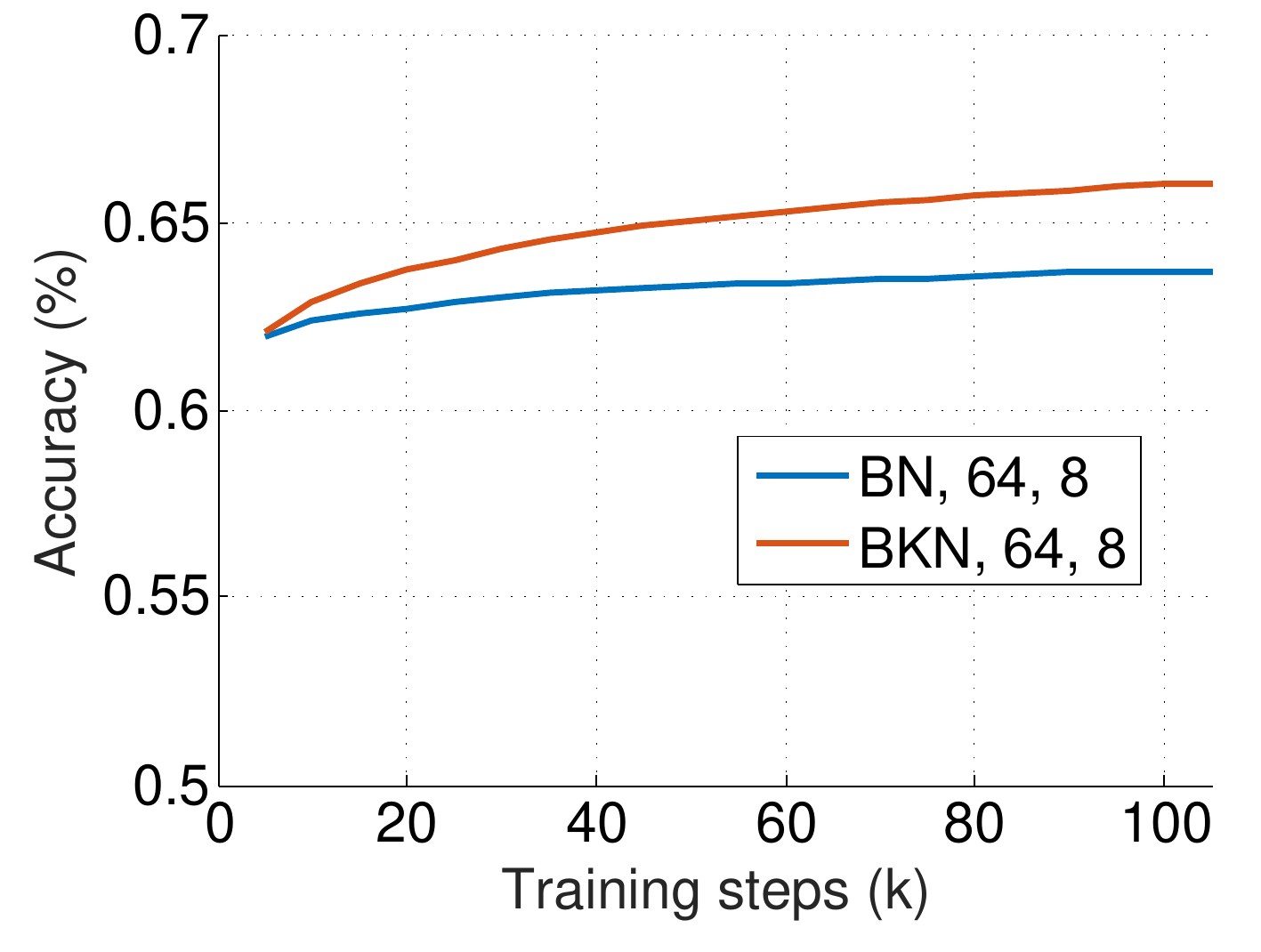}
  \caption{Evaluation on Inceptionv2}
\end{subfigure}%
\begin{subfigure}{0.5\linewidth}
  \centering
  \includegraphics[width=0.99\linewidth]{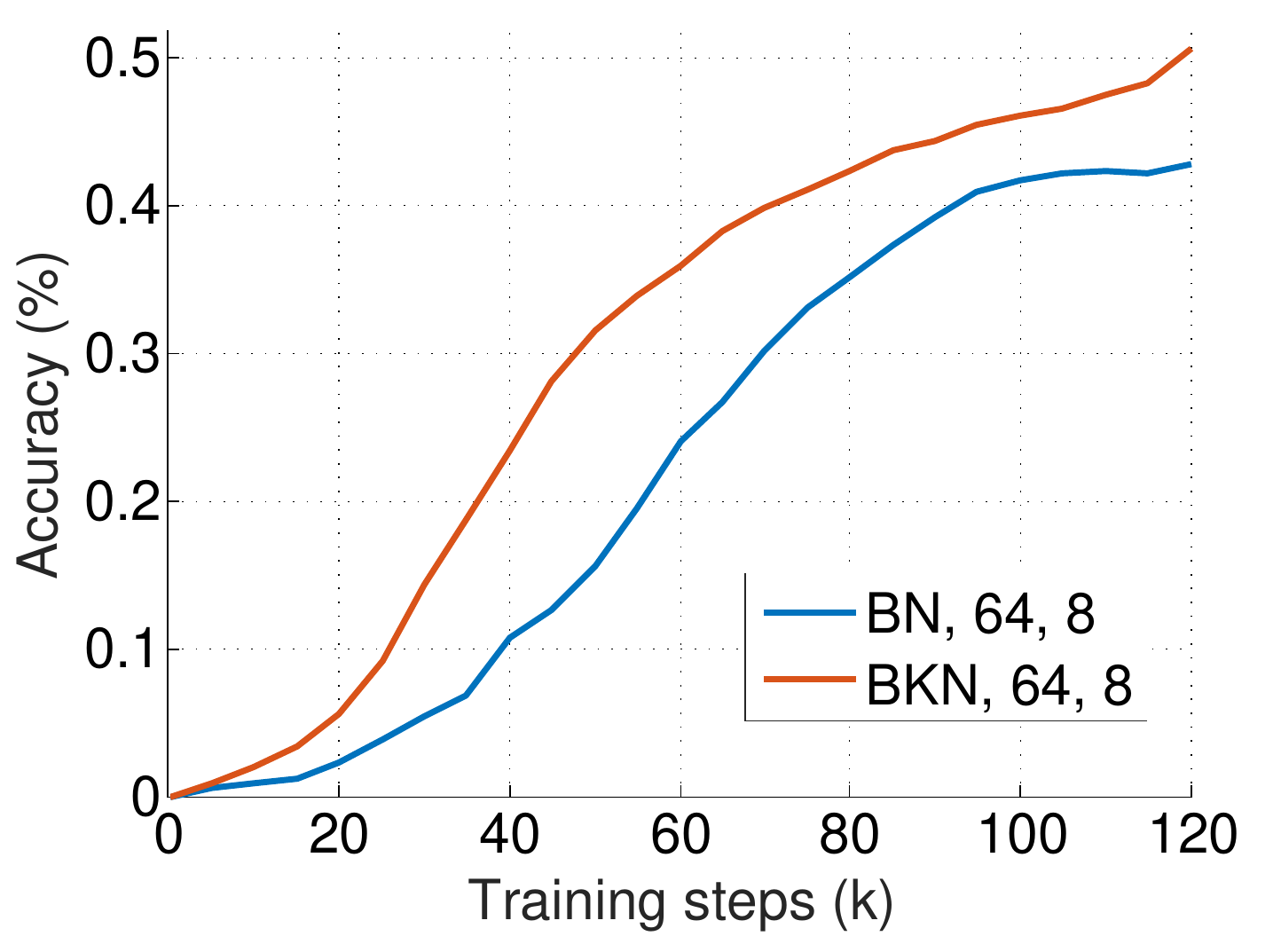}
  \caption{Evaluation on RestNet}
\end{subfigure}
\caption{{Validation accuracy for models trained with BN, and BKN, where the setting is (64, 8). BKN achieves a higher accuracy.}}
\label{fig:64div8}
\end{figure}

\vspace{3mm}
\noindent
\textbf{Setting of (256,1).} This setting is extremely challenging, because normalization is accomplished within each training sample. It is adopted when the input image has massive pixels or each sample is a video containing multiple frames. We have shown that BKN works well in micro-batches as shown in Fig.\ref{fig:256div4}. Next we investigate the necessary of estimating the distribution when \emph{statistic-batch-size} degrades to $1$. In Fig.\ref{fig:256div1} (a) and (b), we compare two options including `A': the batch sample statistics are used to normalize the layer's inputs
(Fig.\ref{fig:256div1} (a)) during inference, and `B': population statistics are used for normalization (Fig.\ref{fig:256div1} (b)) during inference.

We have observe two major phenomenons in Figure \ref{fig:256div1} (a) and (b). First, in both options BKN are significantly better than other methods. In Option `A', BKN obtains a 2.11\% and 2.75\% increase compared to BN and BRN, respectively.

Second, in comparison, option `A' is significantly better than `B'. For example, BKN obtains a top-1 accuracy of 47.99\% in option `A', while 0.4\% in option `B'.
Note that this gain is solely due to the usage of different statistics. We attribute this to \keze{the following two} reasons. First, all approaches fail to estimate the population statistics for 1-example-batch training. As is discussed in Section \ref{sec:3}, the networks are trained using batch sample statistics, while tested based on population statistics obtained by moving averages. This is natural when the \emph{statistic-batch-size} is larger than $1$. But things change when the \emph{statistic-batch-size} equals $1$, where in every iteration the statistics of only $1$ example are calculated and stored to the moving averages.
The information communication never happens between any two examples.
Therefore the moving averages are difficult to represent the population statistics. In fact, this shows that learning a model to approximate the population statistics is infeasible. One possible solution is to use the moving averages to normalize the layer inputs during training, but turns out to be infeasible in \cite{ioffe2015batch, ioffe2017batch}.
Second, we indeed do not need any population statistic in the case of 1-example batches. Recall that the moving averages (population statistics) are introduced based on the following reasons. If there is no population statistic, in order to evaluate an example in test, we need to calculate the statistics online either based on this unique example, or using temporary partners (\ie the other testing samples) to help. Statistics based on single example is not reliable. Using temporary partners suffers the problem of uncertain results when the temporary partners changes. All the above problems disappear when the \emph{statistic-batch-size} equals $1$ because it ensures that the activations computed in the forward pass of training step depend only on a single example. Also note that BKN has a better performance than other methods, improving 2.11\% and 2.75\% compared to BN and BRN respectively. 
These results verify the effectiveness of BKN.

\subsection{Small gradient-batch-size}

In the last three settings, we compare different normalization methods when the \emph{gradient-batch-size} is small.

\vspace{3mm}
\noindent
\textbf{Setting of (64,8).} We have shown that BKN works better than others in micro-batch setting, i.e. its \emph{statistics-batch-size} is typically smaller than 10, leading to nonnegligible optimization difficulty. Next we investigate a more challenging setting, i.e. both its \emph{gradient-batch-size} and \emph{statistics-batch-size} are tiny, indicating that not only the statistics of normalization are estimated by using a few samples, but also the gradients are calculated by using a small minibatch. We first compare the options of (64, 8).

We have two major observations from Figure \ref{fig:64div8}. First, BKN can achieve comparable performance to the baseline using 4 times less \emph{gradient-batch-size} and 4 times less \emph{statistics-batch-size}, while BN cannot reach such performance even using more iterations. For example, in Inceptionv2 as shown in Figure \ref{fig:64div8} (a), \keze{Inception+BKN} achieves 66.02\% top-1 accuracy, while Inception+BN obtains only 63.66\%. Similarly, \keze{ResNet+BKN obtains a 7.81\% improvement gain}, which is a 18.2\% relative improvements.

Second, the performances of all normalization methods in the setting of (64,8) are lower than those in the setting of (256,4). However, BKN suffers less than BN. This indicates that BN has limitation when solving the training problem with micro-batches. On the contrary, BKN provides reasonable results in both settings.

\begin{figure}
\centering
\begin{subfigure}{0.5\linewidth}
  \centering
  \includegraphics[width=0.99\linewidth]{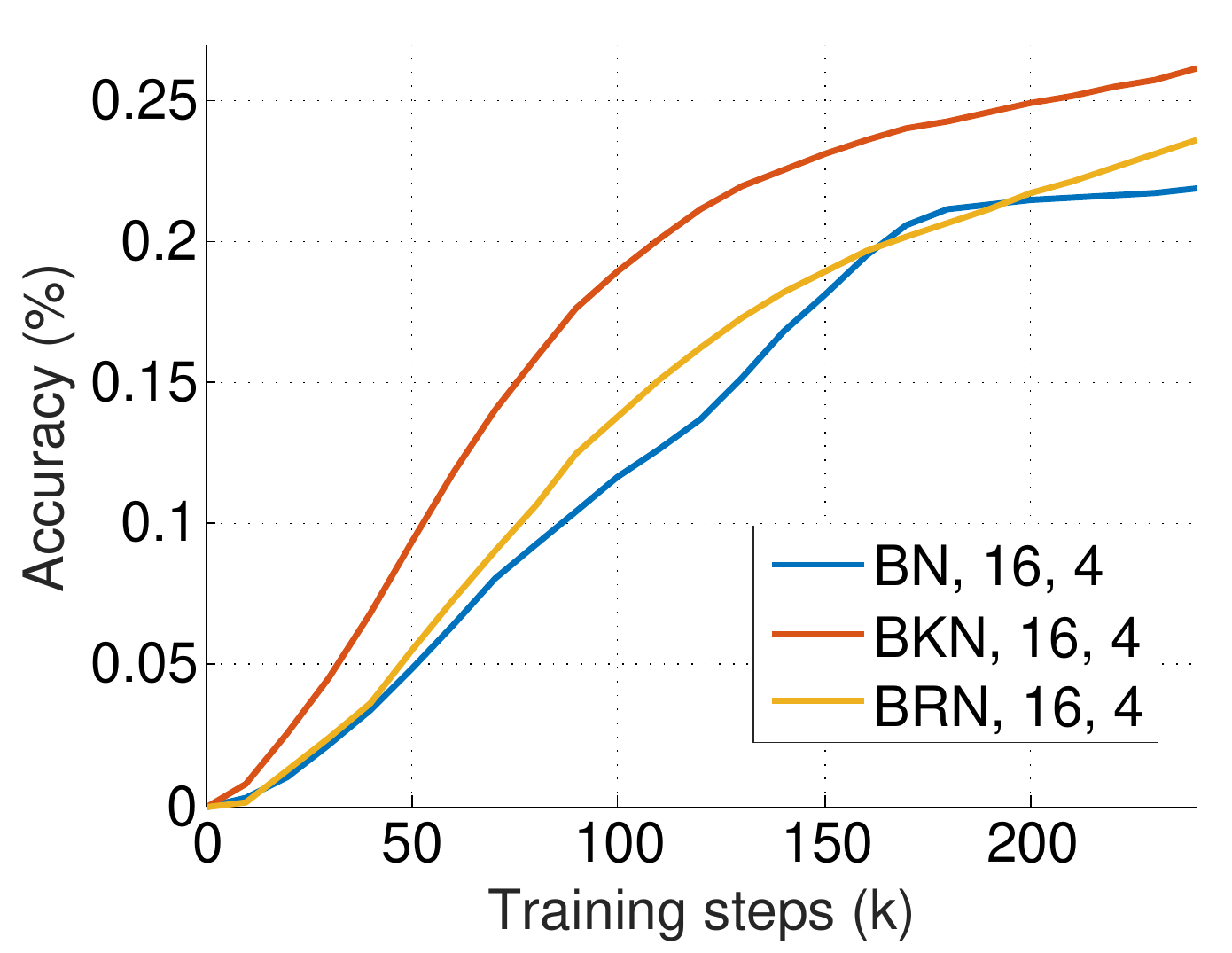}
  \caption{Evaluation on Inceptionv2}
\end{subfigure}%
\begin{subfigure}{0.5\linewidth}
  \centering
  \includegraphics[width=0.99\linewidth]{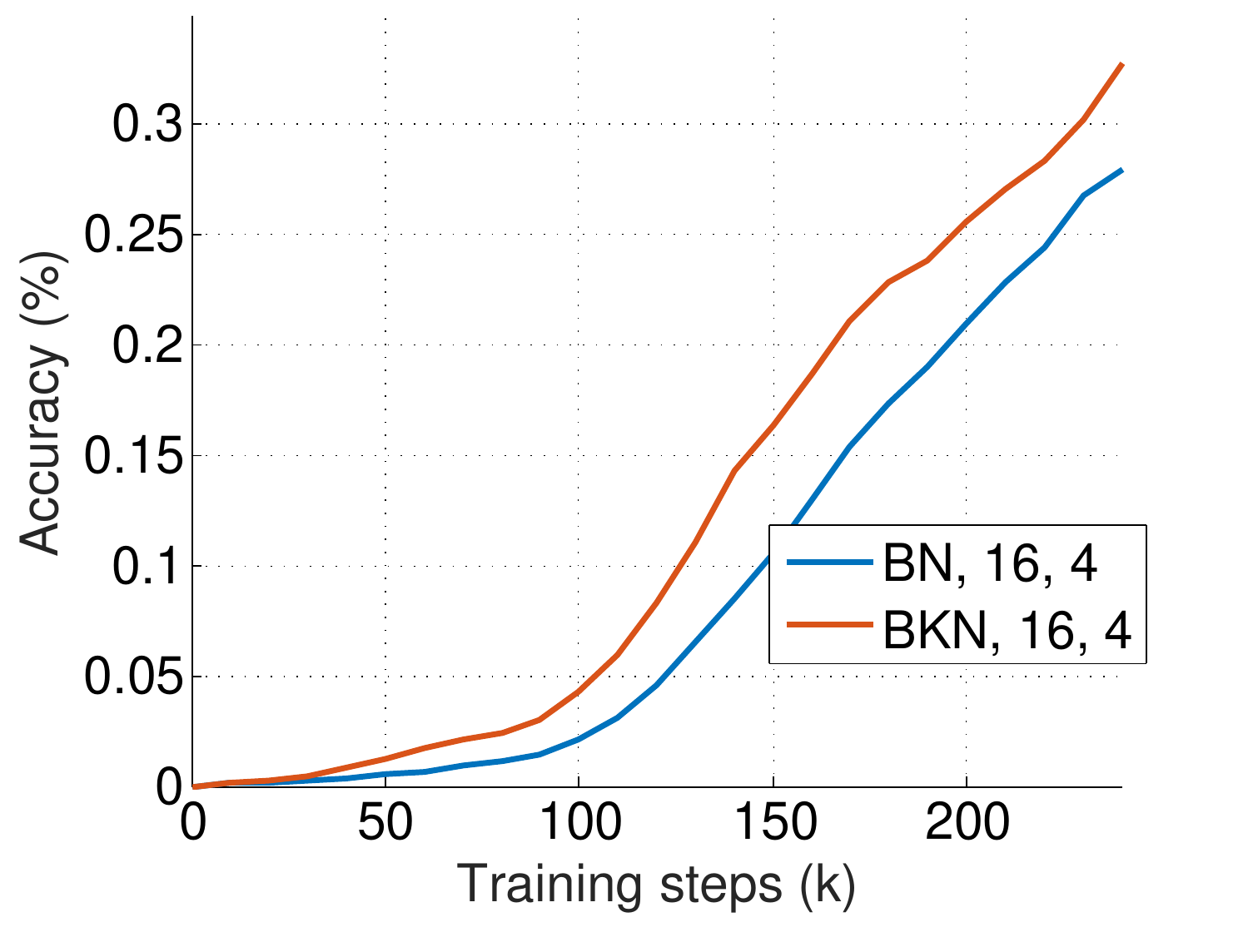}
  \caption{Evaluation on ResNet}
\end{subfigure}
\caption{Validation accuracy for models trained with BN, BRN and BKN, the setting is (16, 4). BKN achieves a higher accuracy.}
\label{fig:16div4}
\end{figure}

\begin{figure}
\centering
\begin{subfigure}{0.5\linewidth}
  \centering
  \includegraphics[width=0.99\linewidth]{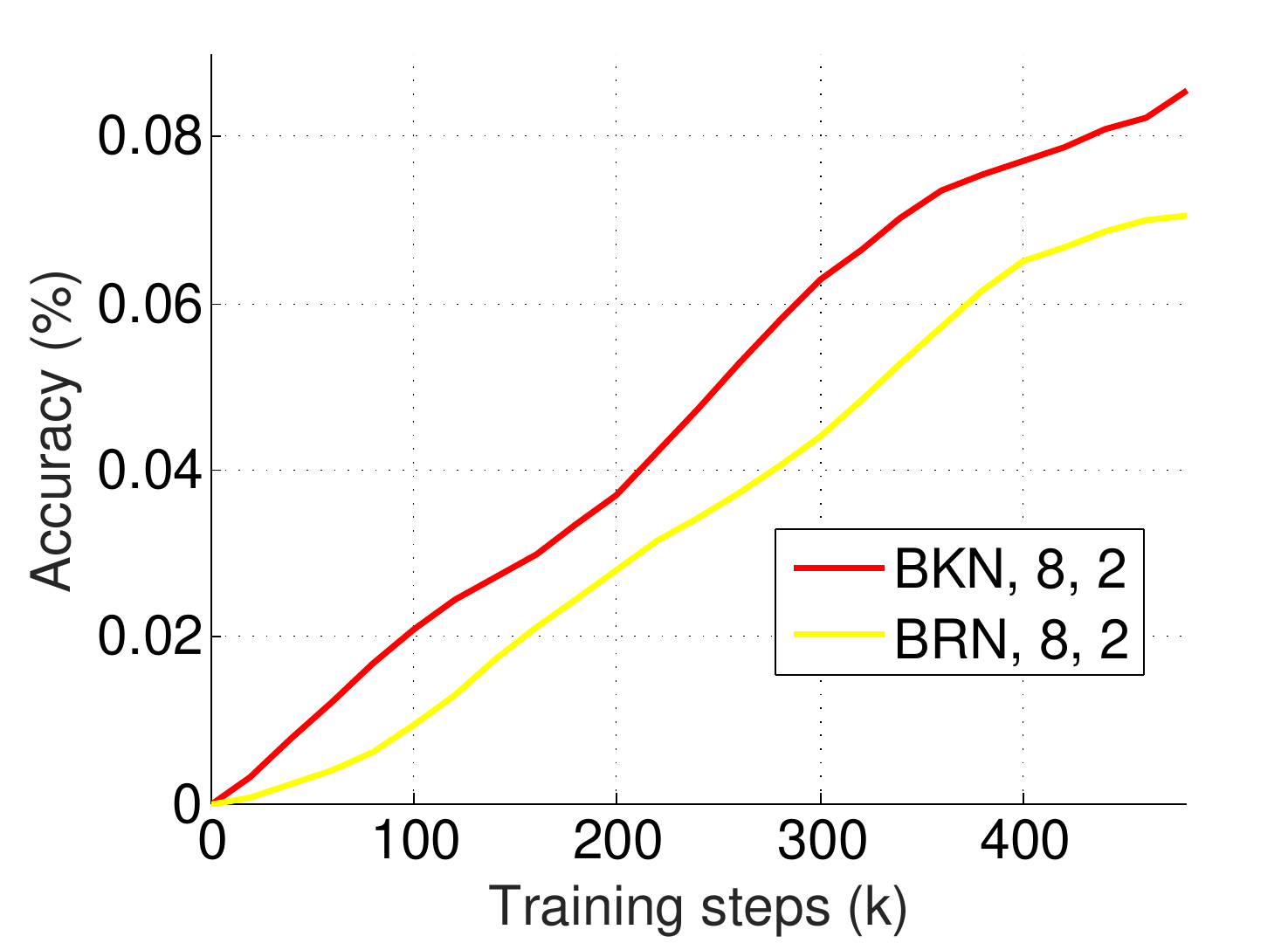}
  \caption{Evaluation on Inceptionv2}
\end{subfigure}%
\begin{subfigure}{0.5\linewidth}
  \centering
  \includegraphics[width=0.99\linewidth]{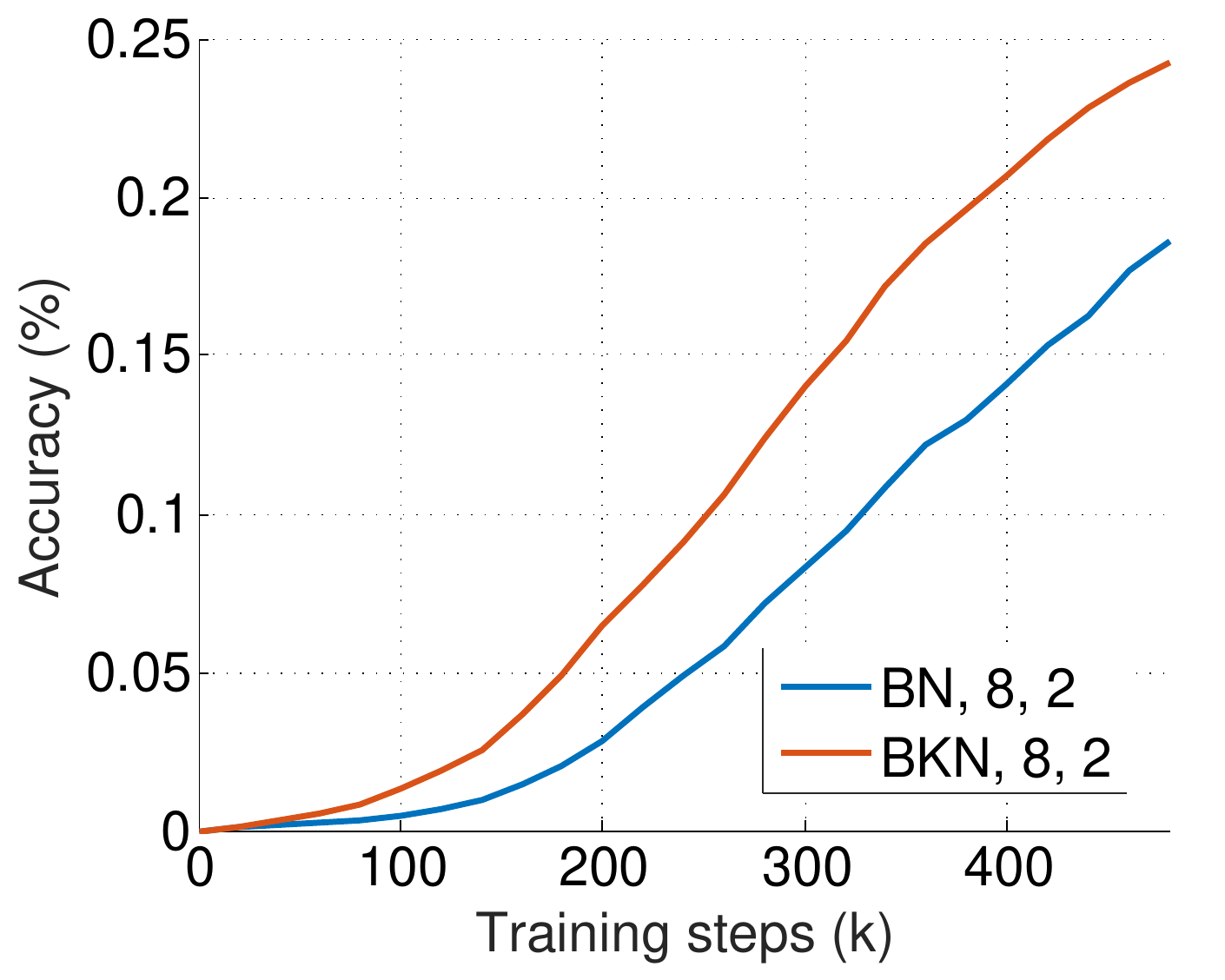}
  \caption{Evaluation on ResNet}
\end{subfigure}
\caption{Validation accuracy for models trained with BN, BRN and BKN, the setting is (8, 2). BKN achieves a higher accuracy.}
\label{fig:8div2}
\end{figure}

\vspace{3mm}
\noindent
\textbf{Setting of (16,4).} Next we evaluate a more challenging setting, where the \emph{gradient-batch-size} is reduced by 16 times, meanwhile the \emph{statistics-batch-size} is reduced by 8 times. Figure \ref{fig:16div4} presents similar findings that \keze{both} BN and BRN have poor top-1 accuracy than BKN. For example, BKN obtains a 4.24\% improvement compared to BN at the $240k$-th iteration on Inceptionv2, and 4.79\% on ResNet101, which are 19.4\% and 17.1\% relative improvements, respectively.

\vspace{3mm}
\noindent
\textbf{Setting of (8,2).} We further explore another challenging micro-batch setting, where (\emph{gradient-batch-size}, \emph{statistic-batch-size}) = (8, 2). Figure \ref{fig:8div2} shows the behaviors of the BN, BRN, and BKN. Similar phenomena as above can be observed, where networks with BKN converge faster than BN when the mini-batces are micro, verifying BKN's advantage. Also note that the accuracy of all methods are significantly lower than the baseline, which normalized over \emph{statistic-batch-size} of 32, suggesting that such an optimization difficulty is a fundamental problem.

\begin{figure}
\centering
\begin{subfigure}{0.9\linewidth}
\centering
\includegraphics[width=0.99\linewidth]{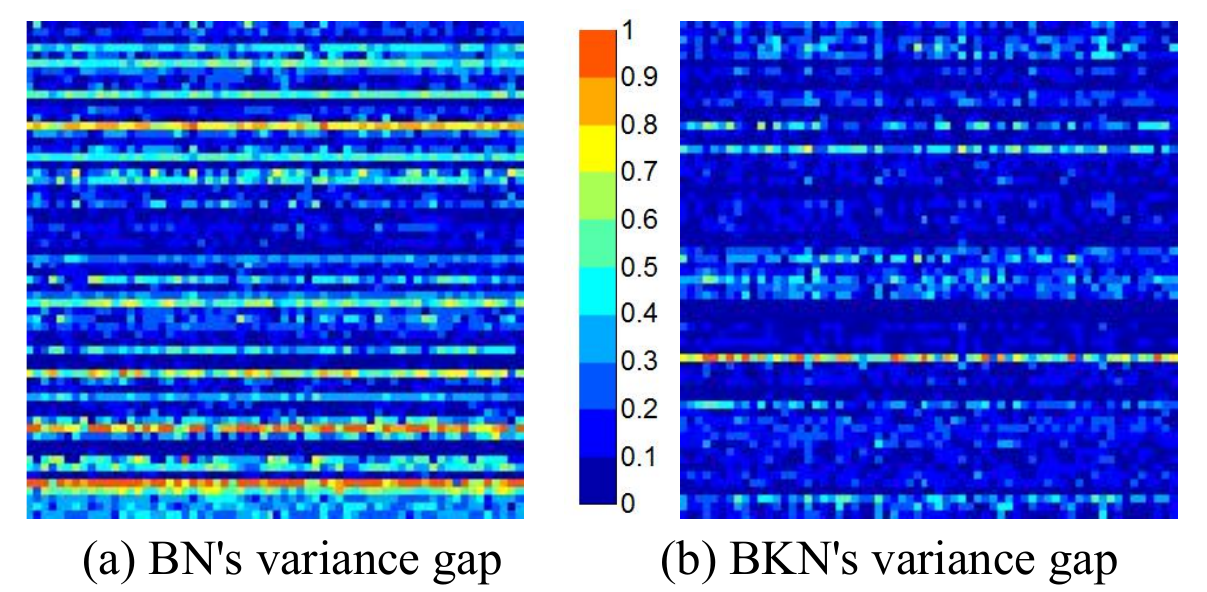}
\end{subfigure}%
\caption{\small{Visualization of variance gap between batch sample variance and moving variance for BN and BKN, respectively, where moving variances are expected to appropriate the population statistics.}}
\label{fig:var_gap}
\end{figure}

\begin{table}
  \caption{\small{Comparison of CIFAR-10/100 \emph{val} accuracy, where `Baseline' is trained using typical batch size of 128, while `BN' and `BKN' are trained using $64\times$ smaller batch size, i.e. $batchsize = 2$.}}
  \scriptsize
  \centering
   \begin{tabular}{l|p{25pt} p{25pt}|p{10pt}p{10pt}|p{18pt}p{18pt}}
    \hline
    & \multicolumn{2}{|c|}{CIFAR-10} & \multicolumn{2}{|c|}{CIFAR-100}&\\
    \hline
    & batch mean/var & moving mean/var & 49M paras & 52M paras & CIFAR-10 & CIFAR-100 \\
    \hline\hline
    BN & 90.0 & 89.4 & 63.8 & 63.1 & 89.4 & 63.8 \\
    BKN & 90.9 & 90.9 &  & 67.3 & 90.9 & 67.3 \\
    Baseline &  & 92.1 & 70.5 & 70.1 & 92.1 & 70.5 \\
    \hline
  \end{tabular}
  \label{tab:estimation}
\end{table}

{\subsection{Analysis on CIFAR-10 and CIFAR-100}}
We \keze{have} conducted more studies on the CIFAR-10 and CIFAR-100 dataset~\cite{krizhevsky2009learning}, both of which consist of 50k training images and 10k testing images in 10 classes and 100 classes, respectively. \keze{Considering our focus is the behaviors of extremely small batch size instead of achieving state-of-the-art results, we design a simple architecture summarized in the following table, where a fully connected layer with 1,000 output channels is omitted.}

\noindent
\vspace{1.5mm}
\begin{scriptsize}
\begin{tabular}{|c|c|c|c|c|c|}
    \hline
    type & conv & inception & inception  & inception & avg pool \\
    \hline\hline
    spatial size & $16\times16$ & $16\times16$ & $16\times16$ & $16\times16$ & $1\times1$  \\
    filters & 32 & 256 & 480 & 512 & 512\\
    $1\times$1 &  & 64 & 128 & 192 &  \\
    $1\times$1/$3\times$3 &  & 96, 128 & 128, 192 & 96, 208&  \\
    $1\times$1/$5\times$5 &  & 16, 32 & 32, 96 & 16, 48 &  \\
    pool/$1\times1$ &  & 32 & 64 & 64 &  \\
    \hline
\end{tabular}
\end{scriptsize}

\vspace{1.5mm}
\noindent
\textbf{Evidence of more accurate statistic estimations.} To show that our BKN indeed provides a more accurate statistic estimation than BN, we present two evidences as follows: (1) Direct evidence. When the training stage finished, we exhaustively forward-propagated all the samples in CIFAR-10 to obtain their moving statistics and batch sample statistics. The gaps between batch sample variance and the moving variance are visualized in Fig. \ref{fig:var_gap} (a) and (b) for BN and BKN, respectively.
In Fig. \ref{fig:var_gap} the horizontal axis represents values of different batches, while vertical axis represents neurons of different channels.
We can observe values in Fig. \ref{fig:var_gap} (b) are smaller than Fig. \ref{fig:var_gap} (a), indicating that BKN provides a more accurate statistic estimation. (2) Indirect evidence. During inference, there are two ways to calculate the classification accuracy, i.e. using the moving mean/variance or batch mean/variance . Experimental results in Table \ref{tab:estimation} show that in BKN, using batch mean/variance achieves the same accuracy as using moving mean/variance. While in BN there's a gap between using batch variance and moving variance. This once again proves that our proposed BKN \keze{contributes to} provide more accurate estimations.

\vspace{1mm}
\noindent
\textbf{Ablation of extra parameters.} To show that the gain doesn't come from extra parameters, we provide two evidences. First, the newly introduced parameters are quite few. The parameter numbers of BN and BKN powered networks \keze{are} 49M and 52M, respectively. Second, we also design an experiment to introduce compensatory parameters for BN \keze{empowered} networks for fair comparison, by enlarging its width. Results in Table \ref{tab:estimation} \keze{show} that such operation has no gain.

\vspace{1mm}
\noindent
\textbf{Scalability of BKN.}  We provide experiments on more datasets such as CIFAR-10 and CIFAR-100. Results in Table \ref{tab:estimation} show that BKN beats BN by large margin on these datasets (i.e. 1.5\% on CIFAR-10 and 3.5\% on CIFAR-100). Moreover, we can observe that the performance of the micro-batch (91.0\% when batch size is 2) is very encouraging compared to that of the typical size (92.1\% when batch size is 128).

\section{Conclusion}

This paper presented a novel batch normalization method, called Batch Kalman Normalization (BKN), to normalize the hidden representation of a deep neural network.
Unlike previous methods that normalized each hidden layer independently, BKN treats the entire network as a whole.
The statistics of current layer depends on all its preceding layers, by using a procedure like Kalman filtering.
Extensive experiments suggest that BKN is capable of strengthening several state-of-the-art neural networks by improving their training stability and convergence speed. More importantly, BN can handle the training with mini-batches of very small sizes. In the future work, theoretical analysis will be made to disclose the underlying merits of our proposed BKN.

\section*{Appendix~}

We present the proof of Eqn.(\ref{eq:statistic}) in the following and the gradient derivation of BKN in Eqn.(\ref{eq:bp}).

\textbf{Proof of Eqn.(\ref{eq:statistic}).} Since $u^k$ is merely a Gaussian noise, $u^k\sim \mathcal{N}(0,R)$, the estimated mean and
covariance of $x^{k}=\mathbf{A}^{k}x^{k-1}+u^{k}$
are
\[
\hat\mu{}^{k|k-1}=\mathbf{A}^{k}\hat\mu{}^{k-1|k-1},
\]
 and
\[
\hat\Sigma{}^{k|k-1}=\mathbf{A}^{k}\hat\Sigma{}^{k-1|k-1}(\mathbf{A}^{k})^{T}+R.
\]

Then it is reasonable to assume the true distribution is a mixture
of the current observation and the prediction from the previous layer
\[
x^{k|k}=p^{k}x^{k|k-1}+q^{k}x^{k},
\]
where $q^{k}$ is a gain factor quantifying the repliance on the observation
and $p^{k}+q^{k}=1$. Therefore, the estimation is also a simple weighted
summation of these two,
\[
\hat\mu{}^{k|k}=p^{k}\hat\mu{}^{k|k-1}+q^{k}\bar{x}^{k}.
\]

Similarly,
\[
\mathbb{E}\left(\left(x^{k|k}\right)^{2}\right)=p^{k}\left(\hat\Sigma^{k|k-1}+\left(\hat\mu^{k|k-1}\right)^{2}\right)+q^{k}\left(S^{k}+
\left(\overline{x}^{k}\right)^{2}\right),
\]
where $(\cdot)^2$ denotes the outer product between two vectors to simplify the notations.
And therefore the covariance of the prediction is
\begin{align*}
\hat\Sigma^{k|k} & =\mathbb{E}\left(\left(x^{k|k}\right)^{2}\right)-\left(\{\hat\mu\}^{k|k}\right)^{2}\\
 &=p^{k}\left(\hat\Sigma^{k|k-1}+\left(\hat\mu^{k|k-1}\right)^{2}\right)+q^{k}\left(S^{k}+\left(\overline{x}^{k}\right)^{2}\right)\\
 &~~~~~-(p^{k}\hat\mu{}^{k|k-1}+q^{k}\bar{x}^{k})^{2}\\
 &=p^{k}\hat\Sigma^{k|k-1}+q^{k}S^{k}+p^{k}q^{k}\left(\bar{x}^{k}-\hat\mu^{k|k-1}\right)^{2}.
\end{align*}

\textbf{Gradient Computations of BKN.}
\begin{small}
\begin{equation} \label{eq:bp}
\small
\begin{aligned}
\frac{{\partial l}}{{\partial {{{{\hat{ x} }}}_i}}} &= \frac{{\partial l}}{{\partial {{{y}}_i}}}\gamma \\
\frac{{\partial l}}{{\partial {{\hat{ \Sigma } }^{k|k}}}} &= \sum\limits_{i = 1}^m {\frac{{\partial l}}{\partial {{\hat{ x} }_i}}} ({x_i} - {{\hat{ \mu } }^{k|k}})(-\frac{1}{2}){{({{\hat{ \Sigma } }^{k|k}})}^{-\frac{3}{2}}}\\
\frac{{\partial l}}{{\partial {{\hat{ \mu } }^{k|k}}}} &= \sum\limits_{i = 1}^m {\frac{{\partial l}}{{\partial {{\hat{ x} }_i}}}} ( - 1){{({{\hat{ \Sigma } }^{k|k}})^{-\frac{1}{2}}}}\\
\frac{{\partial {{\hat{ \Sigma } }^{k|k}}}}{{\partial {x_i}}} &= \frac{2{q^k}}{{m}} ({x_i -q{\bar x}^k - p {\hat \mu}^{k|k-1}} ) \\
\frac{{\partial l}}{{{x_i}}} &= \frac{{\partial l}}{{\partial {{\hat{ x} }_i}}}\frac{1}{\sqrt{{{\hat{ \Sigma } }^{k|k}}}} + \frac{{\partial l}}{{\partial{{\hat{ \Sigma } }^{k|k}}}}\frac{{\partial {{\hat{ \Sigma } }^{k|k}}}}{{\partial {x_i}}}+ \frac{{\partial l}}{{\partial {{\hat{\mu } }^{k|k}}}}\frac{{{q^k}}}{m} \\
\frac{{{\partial{{\hat{ \Sigma } }^{k|k}}}}}{\partial q} &= S^k - {{\hat{ \Sigma } }^{k|k-1}} + (1 -2q^k)({\bar x}^k - {\hat{\mu}}^{k|k-1})^2 \\
\frac{\partial l}{\partial q} &= \frac{{\partial l}}{{\partial{{\hat{ \Sigma } }^{k|k}}}}\frac{\partial{{\hat{ \Sigma } }^{k|k}}}{\partial q} - \frac{\partial l}{\partial {{\hat \mu}^{k|k}}}{{\hat \mu}^{k|k-1}} \\
\frac{{\partial l}}{\gamma } &= \sum\limits_{i = 1}^m {\frac{{\partial l}}{{\partial {y_i}}}{{\hat{ x} }_i}} \\
\frac{{\partial l}}{\beta } &= \sum\limits_{i = 1}^m {\frac{{\partial l}}{{\partial {y_i}}}}
\end{aligned}
\end{equation}
\end{small}

{\small
\bibliographystyle{ieee}
\bibliography{egbib}
}

\end{document}